\documentclass[sigconf, natbib=true]{acmart}
\usepackage{times}
\usepackage{graphicx}
\usepackage{amsthm,amsmath}
\usepackage{booktabs}
\usepackage{subfigure}
\usepackage{color}
\usepackage{mathrsfs}
\usepackage{bm}
\usepackage{multirow}
\usepackage{tabularx}
\usepackage[noend]{algpseudocode}
\usepackage{array}
\usepackage{enumitem}

\usepackage{pdftexcmds}
\usepackage{catchfile}
\usepackage{ifluatex}
\usepackage{ifplatform}
\usepackage{caption}
\usepackage{color}
\usepackage{bbm}

\newcolumntype{L}[1]{>{\raggedright\let\newline\\\arraybackslash\hspace{0pt}}m{#1}}
\newcolumntype{C}[1]{>{\centering\let\newline  \\\arraybackslash\hspace{0pt}}m{#1}}
\newcolumntype{R}[1]{>{\raggedleft\let\newline \\\arraybackslash\hspace{0pt}}m{#1}}

\usepackage{algorithm} 
\usepackage{algorithmicx}
\usepackage{algpseudocode}
\def\bA{\textbf{A}}

\def\bH{\textbf{H}}
\def\bW{\textbf{W}}
\def\bw{\textbf{w}}

\def\bal{\bm{\alpha}}
\DeclareMathOperator*{\argmin}{arg\,min}

\usepackage{multirow}
\usepackage[normalem]{ulem}
\useunder{\uline}{\ul}{}

\renewcommand\footnotetextcopyrightpermission[1]{} 
\begin{document}
\fancyhead{}
\title{Learn Layer-wise Connections in Graph Neural Networks}


%
%

\author{Lanning Wei$^{1,3}$, 
	Huan Zhao$^1$, 
	Zhiqiang He$^{2,3}$}
\affiliation{\institution{$^1$4Paradigm Inc., China, $^2$Lenovo, China}
	\institution{$^3$Institute of Computing Technology, Chinese Academy of Sciences}
}
\email{weilanning18z@ict.ac.cn;zhaohuan@4paradigm.com;hezq@lenovo.com}

\begin{abstract}
In recent years, Graph Neural Networks (GNNs) have shown superior performance on diverse applications on real-world datasets. To improve the model capacity and alleviate the over-smoothing problem, several methods proposed to incorporate the intermediate layers by layer-wise connections. However, due to the highly diverse graph types, the performance of existing methods vary on diverse graphs, leading to a need for data-specific layer-wise connection methods.
To address this problem, we propose a novel framework LLC (Learn Layer-wise Connections) based on neural architecture search (NAS) to learn adaptive connections among intermediate layers in GNNs.
LLC contains one novel search space which consists of 3 types of blocks and learnable connections, and one differentiable search algorithm to enable the efficient search process.
Extensive experiments on five real-world datasets are
conducted, and the results show that the searched layer-wise connections can not only improve the performance but also alleviate the over-smoothing problem.\footnote{Lanning is a research intern in 4Paradigm. This paper has been accepted by \\ DLG-KDD'21.}
\end{abstract}

%


\maketitle
\section{Introduction}

In recent years, Graph Neural Networks(GNNs) have been widly used due to its promising performance on various graph-based datasets, e.g., chemistry~\cite{gilmer2017neural}, bioinformatics~\cite{ying2018hierarchical}, text categorization~\cite{rousseau2015text}, and recommendation~\cite{xiao2019beyond}. Most GNNs follow a neighborhood aggregation schema, also called message passing \cite{gilmer2017neural}, which learns the embeddings of a node by aggregating the embeddings of its neighborhoods. Representative GNN models are GCN~\cite{kipf2016semi}, GraphSAGE~\cite{hamilton2017inductive}, GAT~\cite{velivckovic2017graph} and GIN \cite{xu2018powerful}.
In practice, multiple GNN layers tend to be stacked to improve the model capacity and thus the final performance.
However, it leads to a chain-like architecture by stacking GNN layers, which only uses the node representations of the previous layer and is proved to provide limited improvements~\cite{li2019deepgcns}. 
Besides, the node representations become indistinguishable along with the network goes deeper, i.e., increasing the stacked layers, which is called the over-smoothing problem~\cite{li2018deeper}.
The problem is that it fails to make full use of the information contained in the intermediate layers, which are important for improving the model capacity and alleviating the over-smoothing problem \cite{xu2018representation,fey2019just,li2019deepgcns}.



\begin{figure}[t]
	\centering
	\includegraphics[width=1.0\linewidth]{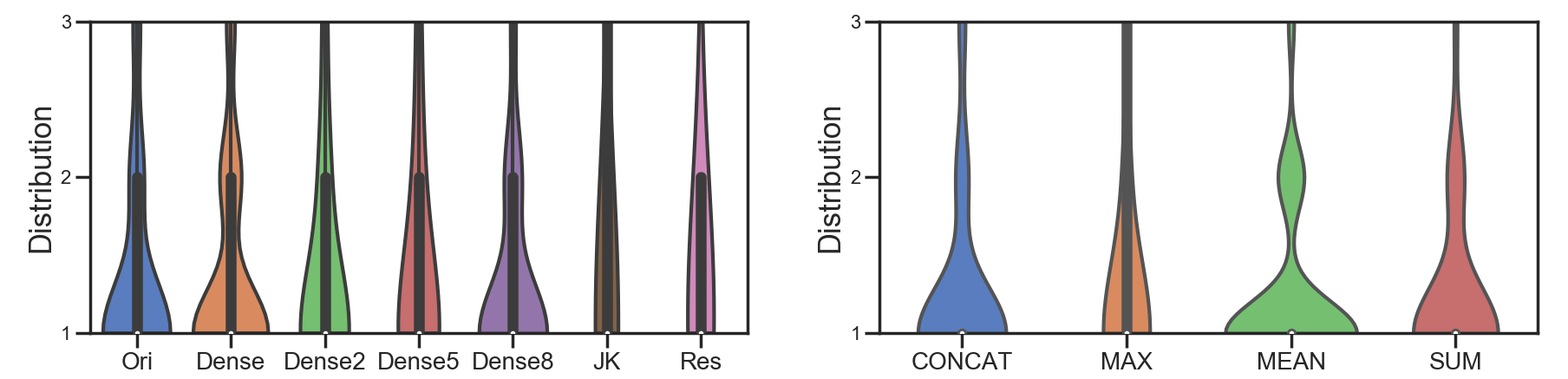}
	\caption{Top 3 rank distributions. Left: The rank distributions of different connection selection methods in GNNs. Right: the rank distributions of different connect functions used in merging these layers.}
	\label{fig-graphgym-top1}
\end{figure}

Recently, to address this problem, several methods are proposed to use the intermediate layers by constructing the layer-wise connections.
For example, JKNet~\cite{xu2018representation} and SIGN~\cite{rossi2020sign} connect all the intermediate layers at the end of GNNs, ResGCN~\cite{li2019deepgcns} summarizes the previous two layers as ResNet~\cite{he2016deep}, and DenseGCN~\cite{li2019deepgcns} concatenates the features of all the previous layers as DenseNet~\cite{huang2017densely}. 
Despite the success of these pre-defined layer-wise connections, in reality, graphs are from highly diverse domains, 
e.g., chemistry~\cite{gilmer2017neural}, bioinformatics~\cite{ying2018hierarchical}, text categorization~\cite{rousseau2015text}, and social networks~\cite{xu2018powerful},  none of these methods can adapt to various tasks.
To further quantify this problem, we design an experiment based on GraphGym~\cite{you2020design}, a GNN benchmark to evaluate the different design dimensions of GNNs, to evaluate existing layer-wise connection methods. As shown in Fig. \ref{fig-graphgym-top1}, different connections are selected and fused in each layer.~\footnote{The experiment details can be found in Section~\ref{sec-exp-set}.}
We sample 180 settings on the node classification task, and we can see that no single architecture can win in all cases on 14 diverse datasets (The rank 1 distribution is nearly uniform).
Therefore, it leads to a straightforward need to obtain data-specific layer-wise connections in GNN architecture design. 



%



For this problem, two questions need to be considered: \textbf{how to select the connections and how to fuse these selected previous layers in each intermediate layer}.
To address these problems, we turn to neural architecture search (NAS) in learning data-specific and SOTA architectures in CNNs\cite{zoph2016neural,liu2018darts,xie2018snas} and GNNs~\cite{gao2019graphnas,zhou2019auto,zhao2020simplifying,zhao2021search,li2020autograph,li2021one,cai2021rethinking}.
For example, AutoGraph~\cite{li2020autograph} learns to select the connections in each intermediate layer, SNAG~\cite{zhao2020simplifying} and SANE~\cite{zhao2021search} learn to select and fuse connections at the end of GNNs. However, most of these methods focus on searching for different aggregation functions, while ignoring the layer-wise connections due to the incomplete search space of GNNs.

Thus, in this work we propose one framework LLC (Learn Layer-wise Connections) to learn the connections among layers adaptively. 
Firstly, we design one search space to represent this problem with one Directed Acyclic Graph (DAG), it contains 1 pre-process block, 1 post-process block and several GNN blocks. Apart from the input block, each block can select connections from all the previous blocks and choose one fusion function to merge these inputs. Then, on top of the designed search space, we develop a differentiable search algorithm to make the learning efficient and effective.
The goal of LLC is to improve the model capacity and alleviate the over-smoothing problem by learning layer-wise connections. In this way, our method can be integrated with any existing GNN models.
And experiments on various datasets demonstrate the effectiveness of the searched layer-wise connection methods by LLC.


To summarize, the contributions of this work are as follows:
\begin{itemize}
	\item In this paper, we proposed one framework LLC in learning layer-wise connections to address the model capacity and the over-smoothing problems.
	\item By modeling this problem as a graph neural architecture search problem, we design a novel search space that can cover existing human-designed layer-wise connection methods. On top of the search space,  we develop an efficient search algorithm to obtain data-specific layer-wise connection methods.
	\item  Extensive experiments show that the searched layer-wise connections can not only improve the performance but also alleviate the over-smoothing problem.
\end{itemize}
\noindent\textbf{Notations}
We represent a graph as $G =(\bA, \bH) $ ,where $\textbf{A} \in \mathbb{R}^{N \times N}$ is the adjacency matrix of this graph and $\bH \in \mathbb{R}^{N \times d}$ is the node features.  $N$ is the node number.
$\widetilde{\mathcal{N}}(v) =  \{v\} \cup \{ u | \textbf{A}_{uv} \ne 0 \} $ represents set of the self-contained first-order neigbors of node $v$, and $\textbf{h}_v^l$ denotes the features of node $v$ in $l$-th layer.

\section{Related Work}
%
%
%

\subsection{Layer-wise Connections in GNNs}
General GNNs are constructed by stacking aggregators and each layer is connected with the previous layer merely. 
It leads to the over-smoothing and performance drop problem on deep networks.
To alleviate the over-smoothing problem and improve the model capacity, layer-wise connections are used in GNNs.
DenseGCN~\cite{li2019deepgcns} concatenates all the previous layers, and ResGCN~\cite{li2019deepgcns} summarizes the previous two layers. These 2 methods utilize connections in each layer, and more methods focus on fuse all the intermediate layers at the end of GNNs directly.
JKNet~\cite{xu2018representation} integrates all the intermediate layers with concatenation, LSTM and maximum; SIGN~\cite{rossi2020sign} uses different adjacency matrix $\bA^n$ in each layer and fuses these layers with concatenation.
%
These pre-defined connections cannot handle the diverse graph data as shown in Figure~\ref{fig-graphgym-top1}. However, our method LLC can learn layer-wise connections adaptively and improve the model capacity as well as alleviate the over-smoothing problem.




\subsection{Graph Neural Architecture Search}
NAS methods were proposed to automatically find SOTA CNN architectures in a pre-defined search space 
and representative methods are~\cite{zoph2016neural,real2017large,pham2018efficient,liu2018darts,xie2018snas,real2019regularized}. Very recently, researchers tried to automatically design GNN architectures by NAS, and there are several pioneering works. 
Based on Reinforcement Learning (RL), which sample architectures with RNN controller and updated with policy gradient~\cite{zoph2016neural,pham2018efficient}, GraphNAS~\cite{gao2019graphnas} and Auto-GNN~\cite{zhou2019auto} learn to design aggregation operations with attention function, attention head number, embedding size, etc.; SNAG~\cite{zhao2020simplifying} provides extra connections selection and layer aggregations learning in the output node. 
Based on Evolutionary Algorithm (EA), which select parent architecture from the search space and generate new architectures with mutation~\cite{real2017large} and crossover~\cite{real2019regularized}, AutoGraph~\cite{li2020autograph} learns to select connections in each intermediate layers.
These RL and EA based methods need thousands of architecture evaluations which are time-consuming and computationally expensive.
Differentiable methods~\cite{liu2018darts,xie2018snas} construct an over-parameterized network (supernet) and optimize this supernet with gradient descent due to the continuous relaxation of the search space. DSS~\cite{li2021one} and  GNAS~\cite{cai2021rethinking} learn to design aggregators in each layer. SANE~\cite{zhao2021search}, the first method to apply differentiable NAS in design GNNs, provide the aggregator learning and extra skip connections and layer aggregations learning for the output node.

More graph neural architecture search methods can be found in~\cite{zhang2021automated}. Compared with these methods focusing on searching for different aggregation functions, our method LLC try to search for data-specific layer-wise connections.
\section{Methods}

\subsection{Overview}
In this section, we elaborate on the proposed framework LLC, which can solve the layer-wise connection learning problem in GNNs. It contains the designed search space and the differentiable search algorithm.

\begin{figure}[t]
	\centering
	\includegraphics[width=1.0\linewidth]{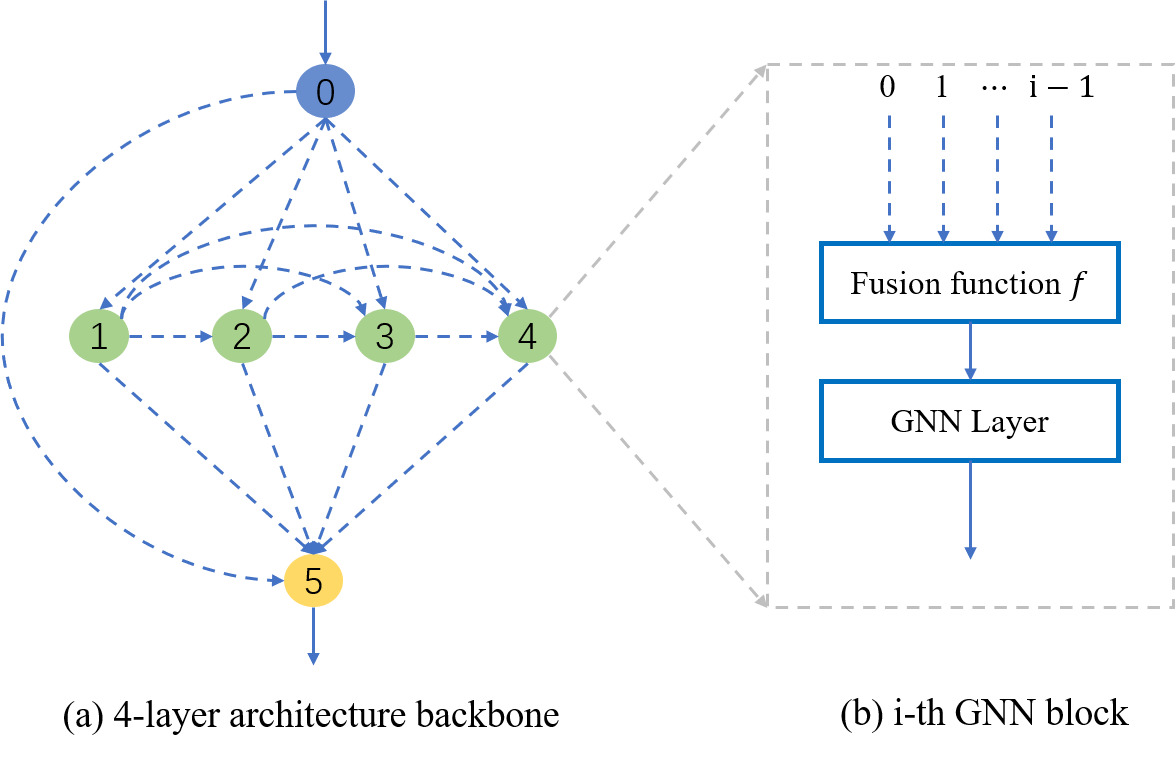}
	\caption{The designed search space. (a) We use a 4-layer architecture backbone as an example, one input block 0, one output block 5 and 4 GNN blocks. The connections among these blocks are learnable. (b) In $i$-th GNN block, the previous layers are selected and then merged with a fusion operation $f$, then the results are used by the following GNN layer.}
	\label{fig-search-space}
\end{figure}

We design one novel search space with the DAG in Fig. \ref{fig-search-space}(a) to learn connections, and it contains
one input block, one output block and several GNN blocks. Based on the learnable layer-wise connections which denoted as dashed edges, the blocks select and fuse connections with as shown in Figure \ref{fig-search-space}(b).
After learning finished, we can construct the GNN architectures by connecting blocks with the learned connections. 
To learn more flexible GNNs rather than chain-structure GNNs stacking message passing layers, the connection from the previous block is learnable either. With this search space, more flexible architectures can be obtained with LLC.


\subsection{Search Space}

We use a DAG to represent the search space as shown in Fig. \ref{fig-search-space}(a), 
the dashed edges represent the learnable connections, there exist 2 candidate operations \texttt{IDENTITY} and \texttt{ZERO} in each dashed edge which corresponding to the selection and not selection stage. 
With the input graph $G=(\bA^I, \bH^I)$, block 0 processes the input features $\bH^I$ with a 2-layer MLPs (Multilayer Perceptrons), and the output of block 0 can be represented as $\bH^0 = \sigma(\bW_1\sigma(\bW_0\bH^I))$. 
GNN blocks (block $1, 2, 3, 4$ in Figure~\ref{fig-search-space}(a)) select and fuse these connections firstly, then update node embeddings with GNN layers as shown in Figure\ref{fig-search-space} (b).
Block 5 is the post-process block, it selects and fuses connections as the GNN block, and then generates the output results with a 2-layer MLPs. 

In this paper, based on the aggregators which have been explored extensively, we learn to select and fuse connections with the selection and fusion operations as shown in Table~\ref{tb-search-space}. These fusion operations produce one new feature based on these connected layers with the summation, average, maximum, concatenation, LSTM cell and attention mechanism, which demonstrated as \texttt{SUM}, \texttt{MEAN}, \texttt{MAX}, \texttt{CONCAT}, \texttt{LSTM} and \texttt{ATT}, respectively.

\noindent\textbf{Discussions.} With the design of this search space, it is capable to obtain the multi-branches architectures  as~\cite{rossi2020sign,abu2019mixhop}, which can expand the chain-structure in general GNNs. Therefore, LLC is more flexible than~\cite{zhao2020simplifying,zhao2021search,li2020autograph} which search connections based on the chain-structure GNNs.

\begin{table}[]
	\caption{The selection set $\mathcal{O}_e$ and the fusion set $\mathcal{O}_f$ we used in LLC.}
	\begin{tabular}{l|l}
		\hline
		& Operations            \\ \hline
		Selection set $\mathcal{O}_e$ & \texttt{ZERO}, \texttt{IDENTITY} \\ \hline
		Fusion set $\mathcal{O}_f$&\texttt{SUM}, \texttt{MEAN}, \texttt{MAX}, \texttt{CONCAT}, \texttt{LSTM}, \texttt{ATT}  \\ \hline
	\end{tabular}
\label{tb-search-space}
\end{table}

\subsection{Differentiable Search Algorithm}

%
\noindent\textbf{Continuous relaxation.}
The search algorithm is used to search architectures from the search space shown in Figure~\ref{fig-search-space}.
To enable the usage of gradient descent to accelerate the search process, continuous relaxation is used to make the search space continuous, thus the discrete selection of operations is relaxed by a weighted summation of all possible operations as \begin{align}
	\label{eq-supernet-weightedsum}
	\bar{o}(x)=\sum\nolimits_{k=1}^{\left|\mathcal{O}\right|} c_ko_k(x),
\end{align}
$c_k \in (0,1)$ is the weight of $k$-th operation $o_k(\cdot)$, it is generated by one relaxation function $c_k = g(\bm{\alpha})$ and $\alpha_k$ is the corresponding learnable supernet parameter for $c_k$. 

For block $j$, the connection selection result from block $i$ can be represented as the weighted summation of 2 selection operations as shown in    
\begin{align}
	\label{eq-block-output}
	\bar{o}^{ij}(\textbf{x}_i)=\sum\nolimits_{k=1}^{\left|\mathcal{O}_e\right|} c_k^{ij}o_k^{ij}(\textbf{x}_i) = c_1^{ij}\textbf{0} + c_2^{ij}\textbf{x}_i=c_2^{ij}\textbf{x}_i.
\end{align} 
These selection results are integrated with fusion operations as shown in
\begin{align}
	\label{eq-combination}
	\bar{o}^j(\textbf{x}) = \sum\nolimits_{k=1}^{\left|\mathcal{O}_f\right|} c_k^{j}o_k^j(\textbf{x}) = \sum\nolimits_{k=1}^{\left|\mathcal{O}_f\right|} c_k^{j}f_k^j(\{\bar{o}^{ij}(\textbf{x}_i)|i=0,\cdots,j-1\}),
\end{align}
where $f_k^j$ is the $k$-th candicate fusion operation in block $j$. After learning the layer-wise connections in block $j$, the results $\bar{o}^j(\textbf{x})$ will be utilized by GNN layers or MLPs.

If the \texttt{ZERO} operation is selected, block $j$ can still obtain the features from block $i$. Due to we only have \texttt{ZERO} and \texttt{IDENTITY} in the selection operations, if block $i$ does not select any connections, the input and output for this block should be a zero tensor. 
It has a large influence on layer-wise connection learning since the successors can also obtain the information from block $i$ no matter which operations are selected. 



 
Thus, we use the Gumbel-Softmax~\cite{jang2016categorical} as our relaxation method to alleviate this problem, as shown in 
\begin{align}
	\label{eq-gumble-softmax}
	c_k &=\frac{\exp((\log\bm{\alpha}_k + \textbf{G}_k)/\lambda)}{\sum_{j=1}^{\left| \mathcal{O}\right|} \exp((\log\bm{\alpha}_j + \textbf{G}_j)/\lambda)},
\end{align}
$\textbf{G}_k=-\log(-\log(\textbf{U}_k ))$ is the Gumble random variable, and $\textbf{U}_k$ is a uniform random variable, $\lambda$ is the temperature of softmax. It is designed to approximate discrete distribution in a differentiable manner and shown useful for supernet training in NAS~\cite{xie2018snas,dong2019searching}. By choosing a small $\lambda$, the weight $c_i$  
becomes one-hot, and the results will become a zero tensor if we do not select any connections. That is, the \texttt{IDENTITY} operation will have a smaller influence on the connection learning results when the \texttt{ZERO} operation is chosen.

In this paper, we directly use the Gumble-Softmax as our relaxation function, an alternative way is to add a temperature in Softmax function as $c_i= \frac{\exp(\alpha_i/\lambda)}{\sum\nolimits_{j=1}^{\left|\mathcal{O}\right|} \exp(\alpha_j/\lambda)}$, we will leave the comparisons into the future work.

\noindent\textbf{Optimization with gradient descent.}
Based on the continuous relaxation in Eq.~\eqref{eq-block-output} and Eq.~\eqref{eq-combination}, we can calculate the mixed results step by step as shown in Figure~\ref{fig-search-space}(a). 
After we obtain the final representations from the output block 5, we calculate the cross-entropy loss on the input data and optimize the supernet. Thus, LLC is to solve a bi-level optimization problem as:
\begin{align}
	\min\nolimits_{\bal \in \mathcal{A}} & \;
	\mathcal{L}_{\text{val}} (\bw^*(\bal), \bal),
	\label{eq-nested-nas-opt}
	\\\
	\text{\;s.t.\;} \bw^*(\bal) 
	& = \argmin\nolimits_\bw \mathcal{L}_{\text{tra}}(\bw, \bal),
	\label{eq-nested-nas-opt:2}
\end{align}
where $\mathcal{A}$ represents the search space, $\mathcal{L}_{\text{tra}}$ and $\mathcal{L}_{\text{val}}$ are the training and validation loss, respectively. $\bm{\alpha}$ is the learnable supernet parameter, $\bw$ is the operation parameter, and $\bw^*(\bal)$ represents the corresponding operation parameter after training. In search process, we update $\bm{\alpha}$ and $\bw$ alternately based on the above continuous relaxation. It lead to an efficient search process with gradient descent.
\noindent\textbf{Deriving Process.} After the search process is finished, we preserve the operations with the largest weights in each module, from which we obtain the searched architecture. 

%
%
%

\section{Experiments}

\begin{table*}[t]
	\small
	\caption{Performance comparisons of our method and all baselines. We report the mean test accuracy and the standard deviation with 10 repeat runs. The best results of each base GNN model are in boldface. ``L2'' and ``L4'' means the number of layers of the base GNN architecture, respectively.}
	\begin{tabular}{l|l|c|c|c|c|c}
		\hline
		GNN Layer            & Connection Method       & Cora                          & PubMed                        & DBLP                 & Computer                      & Physics                       \\ \hline
		\multirow{6}{*}{SAGE} 
		& Stacking (L2)      & 0.8609(0.0050)                & 0.8896(0.0029)                & 0.8358(0.0033)       & 0.9114(0.0030)                & 0.9642(0.0011)                \\ \cline{2-7} 
		& Stacking (L4)      & 0.8568(0.0061)                & 0.8823(0.0028)                & 0.8383(0.0032)       & 0.9052(0.0042)                & 0.9597(0.0014)                \\ \cline{2-7} 
		& ResGCN (L4)       & 0.8566(0.0052)                & 0.8899(0.0025)                & 0.8339(0.0030)       & 0.9151(0.0018)         & 0.9631(0.0017)                \\ \cline{2-7} 
		& DenseGCN (L4)     & 0.8668(0.0059)                & 0.8942(0.0027)                & 0.8330(0.0073)       & 0.9074(0.0051)                & 0.9648(0.0014)                \\ \cline{2-7} 
		& JKNet (L4)        & 0.8647(0.0060)                & 0.8921(0.0029)                & 0.8394(0.0062)       & 0.9121(0.0030)                & 0.9656(0.0005)                \\ \cline{2-7} 
		& LLC (L4)          & \textbf{0.8772(0.0050)}          &  \textbf{0.8948(0.0023)}        &  \textbf{0.8440(0.0015)} & \textbf{0.9173(0.0020)}
		& \textbf{0.9662(0.0010)}          \\ \hline
		\multirow{6}{*}{GAT}  
		& Stacking (L2)      & 0.8592(0.0072)                & 0.8756(0.0022)                & 0.8434(0.0026)       & 0.9149(0.0021)                & 0.9576(0.0016)                \\ \cline{2-7} 
		& Stacking (L4)      & 0.8616(0.0055)                & 0.8573(0.0034)                & 0.8429(0.0041)       & 0.8673(0.0874)                & 0.9347(0.0393)                \\ \cline{2-7} 
		& ResGCN (L4)       & 0.8466(0.0092)                & 0.8756(0.0044)                & 0.8411(0.0034)       & 0.8569(0.1618)                & 0.9567(0.0028)                \\ \cline{2-7} 
		& DenseGCN (L4)     & 0.8531(0.0086)                & 0.8867(0.0019)                & 0.8343(0.0037)       & 0.9130(0.0037)                & 0.9616(0.0006)                \\ \cline{2-7} 
		& JKNet (L4)        & 0.8655(0.0046)                & 0.8971(0.0016)                & 0.8373(0.0035)       & 0.9180(0.0023)                & 0.9620(0.0009) \\ \cline{2-7} 
		& LLC (L4)          &\textbf{0.8777(0.0006)} & \textbf{0.8978(0.0023)} & \textbf{0.8490(0.0015)} & \textbf{0.9192(0.0008)} & \textbf{0.9678(0.0002)}                \\ \hline
	\end{tabular}
	\label{tb-performance-agg}
\end{table*}

\subsection{Experimental Settings}
\label{sec-exp-set}
\noindent\textbf{Datasets.} In this part, we evaluate our method on 5 real-world datasets with diverse types and graph size as shown in Table \ref{tb-dataset}. We split the dataset with 60\% for training, 20\% for validation and 20\% for test with the considering of supernet training and evaluation.

\begin{table}[]
	\centering
	\footnotesize
	\caption{Statistics of the 5 real-world datasets in our experiments.}
\begin{tabular}{l|c|c|c|c|c}
	\hline
	& \#Nodes & \#Edges & \#Features & \#Classes & Type          \\ \hline
	Cora     & 2,708   & 5,278   & 1,433      & 7         & Citation      \\ \hline
	PubMed   & 19,717  & 44,324  & 500        & 3         & Citation      \\ \hline
	DBLP     & 17,716  & 105,734 & 1,639      & 4         & Citation      \\ \hline
	Computer & 13,381  & 245,778 & 767        & 10        & Co-purchase   \\ \hline
	Physics  & 34,493  & 495,924 & 8,415      & 5         & Co-authorship \\ \hline
\end{tabular}
\label{tb-dataset}
\end{table}
\noindent\textbf{Baselines.} 
In our experiment, we propose one framework LLC to learn layer-wise connections in GNNs.
We evaluate our method based on GraphSAGE and GAT, 2 widely used methods in learning node representations. We further provide 5 layer-wise connection variants to make comparisons: 2 and 4 layers GNNs without layer-wise connections, 4-layer baselines ResGCN, DenseGCN and JKNet. 

\noindent\textbf{Implementation details.}
We tune these baselines and the searched architectures with the same hyperparameters space and same epoch on HyperOpt~\footnote{https://github.com/hyperopt/hyperopt}. After finetune, we report the test accuracy and the standard deviation with 10 repeat runs.

\noindent\textbf{Setups for Figure \ref{fig-graphgym-top1}.} In the GraphGym experiment, to enable the layer-wise connection learning, we set the message passing layer numbers as [3,4,5,6]. We provide 7 baselines: Ori: which has no layer-wise connections; JK, Res and Dense: which use the same layer-wise connections as  JKNet, ResGCN and DenseGCN; 3 extra baselines Dense3, Dense5 and Dense8: which randomly select the layer-wise connections based on DenseGCN with probability 0.3, 0.5 and 0.8, these 3 baselines are generated before evaluation.


\subsection{Performance Comparisons}


As shown in Table \ref{tb-performance-agg}, there is no absolute winner from 5 baselines on these diverse datasets.  
3 variants using layer-wise connections generally have a better performance than 2 stacking baselines, which indicates the effectiveness of connections among intermediate layers. LLC consistently outperforms all baselines on all datasets, which demonstrates that our method can improve the model performance based on the adaptive layer-wise connections.

\begin{figure}[t]
	\includegraphics[width=0.75\linewidth]{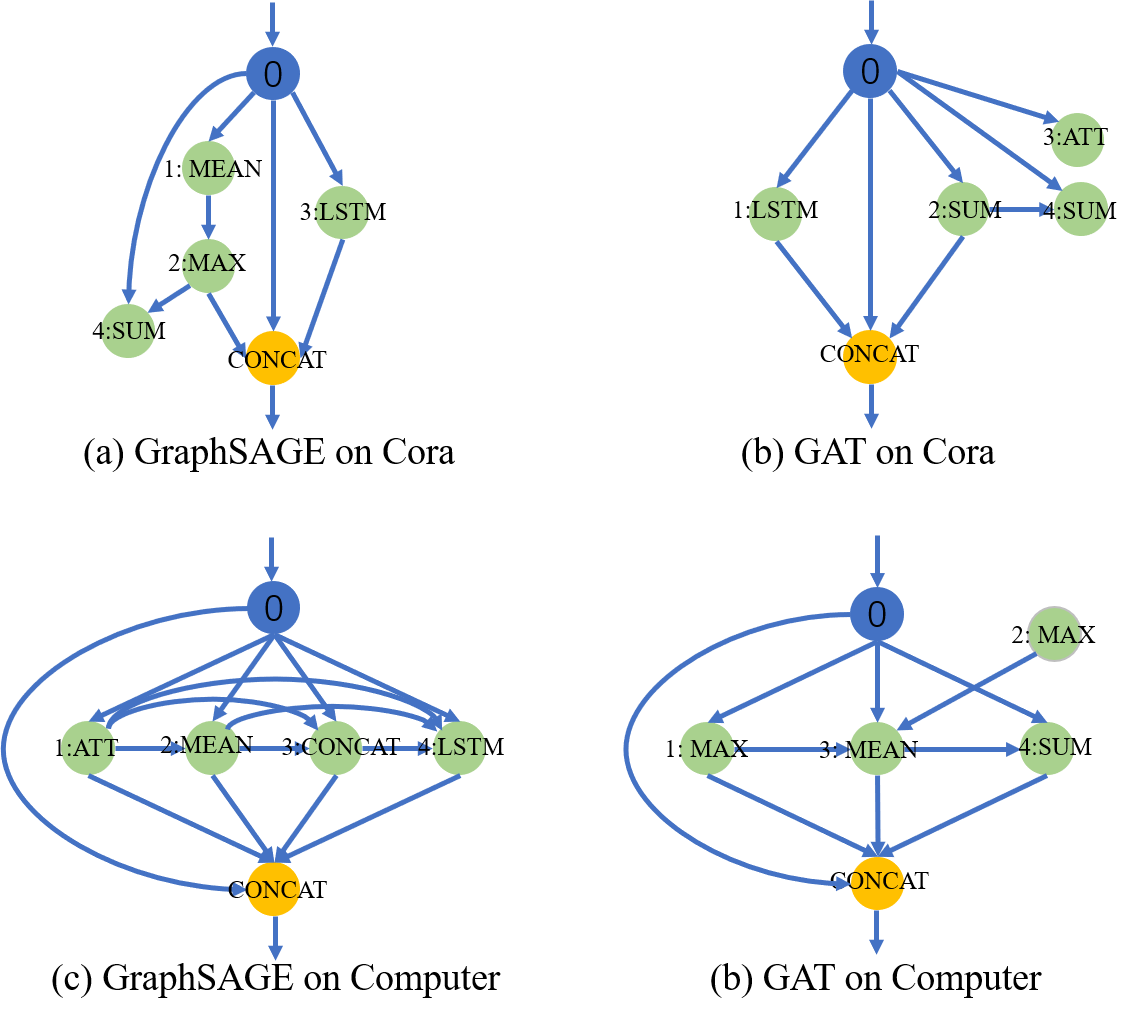}
	\vspace{-6pt}
	\caption{The searched layer-connections on Cora and Computer datasets with 2 base GNN architectures: GraphSAGE and GAT.}
	\label{fig-searched_arch}
\end{figure}

\begin{figure}[t]
	\includegraphics[width=0.7\linewidth]{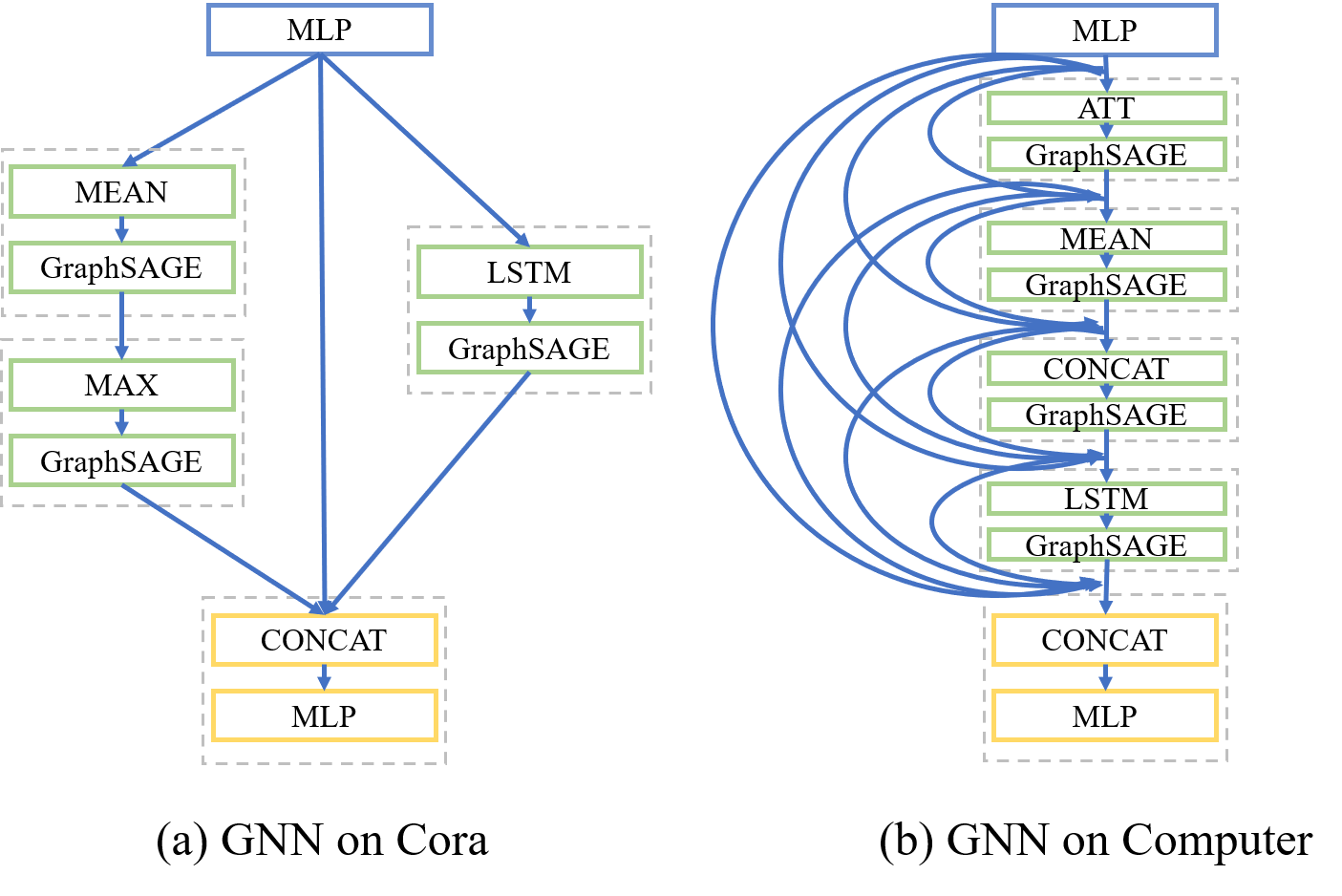}
	\vspace{-6pt}
	\caption{Based on the searched connections in Figure~\ref{fig-searched_arch}, we construct the corresponding architectures with GraphSAGE. On Cora dataset, $4$-th GNN block is removed since it is not used in the output block.}
	\label{fig-constructgnn}
\end{figure}

\textbf{Searched architectures.} We further visualize the searched layer-wise connections in Figure~\ref{fig-searched_arch}. For simplicity, we only show the results on Cora and Computer. Based on the searched connections, the corresponding searched architectures can be represented as Figure~\ref{fig-constructgnn}.
We emphasize several interesting observations in the following:
\begin{itemize}
	\item The results are different for different GNN models on different datasets, demonstrating the necessity of data-specific connection methods.
	\item  On Cora, the searched connections actually lead to multi-branch GNN architectures as shown in Figure~\ref{fig-constructgnn}(a), which is quite different from most existing human-designed GNN architectures, i.e., chain-structure by stacking GNN layers. It provides more flexible and expressive GNN architectures than existing human-designed and searched ones.
	\item More interestingly, the input block $0$ is selected by most blocks, which demonstrates the importance of the input features for the final performance.
	\item On Computer with GraphSAGE, the searched connection method leads to a densely connected GNN architecture similar to the DenseGCN as shown in Figure~\ref{fig-constructgnn}(b). While the performance of LLC on Computer is better than DenseGCN, we attribute the performance gain to the usage of the fusing functions.
\end{itemize}

\subsection{Alleviating the Over-smoothing Problem}

\begin{figure}[t]
	\centering
	\includegraphics[width=0.9\linewidth]{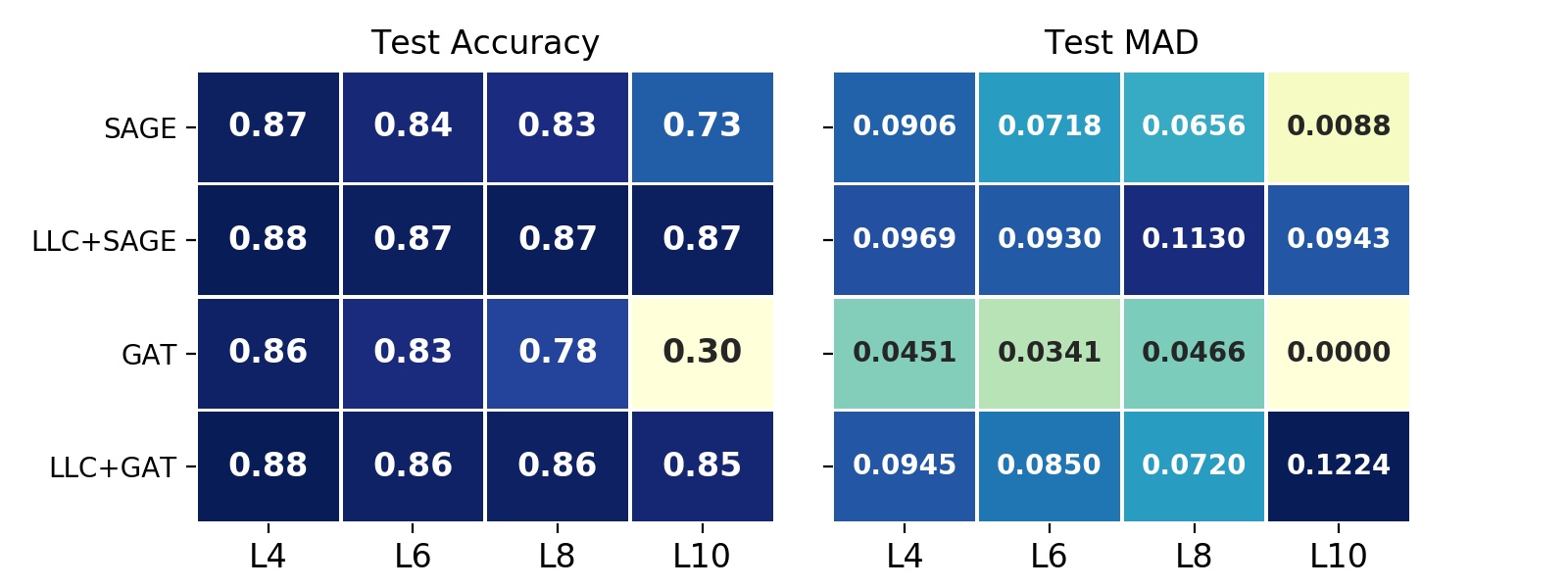}
	\vspace{-6pt}
	\caption{The comparisons of test accuracy and test MAD on different layers, and darker colors mean larger and better values. ``L4'' represents a 4 layer GNN, and so on.}
	\label{fig-mad-layer}
\end{figure}

Along with the increasing layer numbers, connected node pairs have gradually similar receptive fields and lead to similar node representations. That is, node representations become indistinguishable as the network goes deeper. MAD (Metric for Smoothness)~\cite{chen2020measuring} is used to measure the smoothness of the features. We show the comparisons of test accuracy and test MAD in Figure~\ref{fig-mad-layer} on different layers. Along with the layer increase, LLC has less performance drop and the higher MAD values based on baselines, which demonstrates the effectiveness of LLC in alleviating the over-smoothing problem.




\section{Conclusion and Future Work}
In this paper, we propose one framework LLC to address the layer-wise connections learning problem in GNNs.
This framework contains one novel search space which consists of 3 types of blocks and learnable connections, and one differentiable search algorithm to enable an efficient search process. 
The performance has shown that LLC can achieve the SOTA performance with the adaptive layer-wise connections and can alleviate the over-smoothing problem. 
In future work, we will investigate the influence of different aggregation functions and the continuous relaxation methods, and learn in-depth the connections between the searched architectures and the graph properties.

\bibliographystyle{ACM-Reference-Format}
\bibliography{main}

\end{document}